# SYSTEM-GENERATED REQUESTS FOR REWRITING PROPOSALS

*Pietro Speroni di Fenizio[1], Cyril Velikanov[2]*

*We present an online deliberation system using mutual evaluation in order to collaboratively develop solutions. Participants submit their proposals and evaluate each other's proposals; some of them may then be invited by the system to rewrite "problematic" proposals. Two cases are discussed: a proposal supported by many, but not by a given person, who is then invited to rewrite it for making yet more acceptable; and a poorly presented but presumably interesting proposal. The first of these cases has been successfully implemented. Proposals are evaluated along two axes—understandability (or clarity, or, more generally, quality), and agreement. The latter is used by the system to cluster proposals according to their ideas, while the former is used both to present the best proposals on top of their clusters, and to find poorly written proposals candidates for rewriting. These functionalities may be considered as important components of a large scale online deliberation system.*

## 1. Introduction

Our study presented in this paper relates to *open online deliberation* and *open online collaboration*. Within the context of *eParticipation*, online deliberation is indeed an essential aspect; while online collaboration should also be considered as essential if we want eParticipation to be *purposeful*, that is, looking for a solution to a problem stated, and *productive*, that is, yielding result(s) commonly agreed upon.

As eParticipation still remains a rather vague concept, we start with better delimiting the kind of activities we are considering in this paper. Namely, we will consider that a given eParticipation activity (project, campaign) involves a number of *active participants*, i.e. participants who may write their own *contributions*, but who *at least* are actively reading and *appraising* (aka *evaluating*) each other's contributions. A contribution basically is a *proposal* (of how to solve the problem being discussed) or a *comment* on one or more other contributions (proposals or comments). Other types of contributions may indeed be considered as well, but we are mostly interested in proposals, and then, to a lesser extent, in comments.

Hereafter active participants will be simply called *participants*; while those other people who, even after registering, do not manifested their presence, simply won't be considered at all.

### 1.1 Clustering of Contributions

One of our basic assumptions is that an open online deliberation on an important societal issue, when it is expected to be purposeful and productive, may attract a really large number of active participants (in the order of thousands, or tens of thousands or more). In this case, it will be termed a *mass online deliberation* (*MOD*). Within a MOD, a large number of contributions (proposals and comments), all related directly or indirectly to the same problem stated in advance, will be "put on the table" by participants, requiring their reading, appraisal and further commenting. Indeed, an average participant, who devotes for this activity just some part of his/her leisure time, wouldn't be able to read, understand and meaningfully classify such a mass of competing proposals and of various comments on them. Hence the need of a programmed *MOD support system* to facilitate this task for participants.

---

[1] Pietro Speroni di Fenizio (speroni@dei.uc.pt ), CISUC, Department of Informatics Engineering, University of Coimbra, Pólo II, 3030-290 Coimbra, Portugal

[2] Cyril Velikanov (cvelikanov@gmail.com), "Memorial", Malyi Karetnyi per. 12, Moscow 127051, Russia, and PoliTech Institute, 67 Saint Bernard St., Brussels 1060, Belgium

However, while considering the task of semantic sorting (*clustering*) of a large number of proposals/comments on a given issue, we should state that this task cannot be performed automatically by any existing algorithm of automated text analysis—simply because it must analyse not only the *content* of every proposal (what it is about), but also its *intent* (what the author suggests, prefers, wants, or rejects...); for the research into algorithms of the intent analysis of texts is still at its very beginning.

On the other hand, in an open mass deliberation, semantic clustering of proposals cannot be commissioned to a staff of "facilitators" external to the deliberating community—because this would often bias the course of deliberation; and yet more often, if not always, it would arouse suspicion that such a bias is indeed present. We refer to [1] for a more detailed rationale.

Hence, the task of clustering proposals in a deliberation can only be performed collectively by the whole deliberating community, and the total number of items to be considered implies that performance of this task should be distributed among a great number of participants, if not on all of them, acting in a concerted manner. Such a distributed collective action can only be performed under the guidance of a MOD support system with a suitable *clustering algorithm*.

## 1.2  Ranking of Contributions Within a Cluster

The task of clustering proposals should result in grouping together proposals that are similar or compatible in the view of those participants who have read (and appraised) them. Some of those clusters may become rather large, thus necessitating a finer analysis. On the other hand, a given proposal containing some idea may present the idea in a more or less clear, concise and argumented way. On the opposite end of this "quality scale" we may have many poorly formulated proposals, down to barely understandable or completely obscure texts. The latter case taken apart, such a quality ranking, if performed by participants and aggregated by the support system, can further help participants to navigate across a large number of proposals.

As for texts that have appeared obscure to most of their readers, such a text nevertheless may have been sufficiently well understood by some of them, who therefore may have been able to emit a further judgement on the ideas it contains—in a form of a simple (dis)agreement or of a comment. Indeed, it would be very desirable to spread this understanding onto the rest of the deliberating community, for a badly formulated idea may nevertheless be valuable and original. This may need partial or full *rewriting* of a text that appears obscure or is misunderstood.

### 1.3   Collaboration by Rewriting Proposals

Now, let us turn to *online collaboration*—which may be considered either as an integral part of a purposeful online deliberation, or as a separate task. In the former case, it is however hardly imaginable that a very large number of participants may be able to productively collaborate, e.g. in preparing a common "final proposal" from a set of "initial" ones.

A more practical solution would consist in creating within a large deliberating community one or more *online workgroups* of a limited size (say, typically up to 30 people) charged with editing a common proposal. In such a case, as well as in that of a standalone collaborative workgroup, there should be an established mechanism that supports the workgroup's efforts to reach an agreement on a commonly accepted final text, by *rewriting* the input texts, maybe even several times.

Rewriting may be done collectively, or by individual workgroup members; in the latter case, the support system may be helpful in finding out the best candidates for rewriting.

### 1.4   System-Requested User Actions

In each of those above-presented cases, there appears a need for the support system to assist the deliberation and/or collaboration process by requesting specific actions to be performed at specific times by designated participants. These actions may include, in particular: (1) additional appraisal (evaluation) of a given contribution (proposal or comment) that hasn't been sufficiently seen yet, to reliably assign it to some cluster and to rank it within that cluster; (2) rewriting a poorly written proposal that may nevertheless present some interest for the community; (3) rewriting a controversial proposal in order to make it more acceptable to a larger part of the community or of a workgroup.

The idea of system-requested user actions has several aspects, that will be discussed in the following sections. First, we need an algorithm of selecting the most necessary action(s) whose performance is to be system-requested at a given time. Then, we need a principle for selecting the participant(s) the best positioned for successfully performing the requested action(s).

Then, we should consider the motives that would prompt participants to perform the requested action rather than to decline, and whether any incentives should be established to create such motives.

We will discuss these aspects for each of the above contexts (1) – (3), starting from the third.

## 2. Rewriting Proposals by Those Who Disagree With Them

The idea of charging the system with the task of requesting from specific participants to rewrite specific proposals with which they disagree has been implemented by one of the authors in a Web-based collaboration tool "Vilfredo goes to Athens" (http://vilfredo.org). The website is actually open for participation; its description can be found in [2]. In this system, participants can state a problem, and then move their own proposals (on how to solve the problem stated) and evaluate each other's proposal. The system uses some *human based genetic algorithm* (HBGA) [6], where the whole process may have several cycles each consisting of a *proposal phase* and *evaluation phase*. In every consecutive cycle, a new *generation* of proposals is written, based on, or inspired by, a filtered selection of the proposals of the previous generation. The selection is done by extracting a *Pareto front* [7] of the proposals.

The concept of Pareto front can be explained as follows. Let us say that a proposal A *dominates* a proposal B if everybody who votes for B also votes for A, though there exist some participants (at least one) who vote for A but not for B. So we can exclude the proposal B from the list of proposals without making anyone really unhappy because they are still satisfied with one of the available proposals, namely A. The Pareto front then will be the list of proposals that are not dominated and hence pass the filter.

This particular way in which proposals reach the next generation may lead to some singular effects. One of such effects is, that sometimes a certain person has a stronger than average effect on the Pareto front. For example, let us suppose that there are two proposals A and B, and everybody (except John) who voted for B also voted for A, while at the same time there are other people who voted for A but not for B. So A "nearly dominates" B. Everybody except John either prefers A to B, or finds them equally acceptable.

In such a case, easily discoverable by the support system, John can be invited to rewrite A, in a way he thinks would be acceptable for him while (presumably) still acceptable for those who currently accept the original proposal A. If John's version of A, say A', appears to be really acceptable for everybody who voted for A, and if John also accepts it (something we can assume since he wrote it), then A' will dominate both A and B, and we have shrunk the Pareto front at least from this one proposal. In a typical generation we may have multiple people who are invited to rewrite some proposals; moreover, the same proposal can be rewritten by multiple people (each one trying to make the proposal dominate a different one). Also, the same person can also be invited to rewrite multiple proposals. In general, the more a person will vote in a different way from the other people (and thus will represent a unique point of view) the more the system will ask him or her to rewrite other's proposals.

Field experience has shown that participants tend to welcome the invitation to work on a particular proposal. In http://vilfredo.org, at every generation each user has many possibilities. He/she can rewrite another person's proposal, or try to work out a compromise between two different positions, or recover a proposal that was lost, trying to reframe it in a different way. Among all those possibilities it is common for users to feel lost. As time is always limited, a participant will want to make the most effective action. An invitation presented by the support system will represent in this sense one of the most cost-effective possible actions, and is thus generally well received.

## 3. Clustering, Ranking, and System-Requested Appraisal Actions

Our next step concerns a support system for mass online deliberation (MOD), the need for which has been discussed above in the introductory chapter of this paper. This system will include an algorithm of clustering and ranking participants' contributions—proposals, comments, and more—based on individual appraisals (evaluations) of every contribution by a limited number of other participants.

To assure a more objective appraisal of every contribution, esp. immediately after it is written and uploaded into the system, participants who come online will be invited to evaluate proposals that the system presents to them, at random. This should be done in such a way that each proposal is evaluated on average the same number of times, and all are evaluated at least a predefined minimum number of times. Also, at this stage it should not be possible for any participant to see the evaluation results of other ones, to avoid any unconscious bias. If at the end of this "blind evaluation" period some proposals have not yet been evaluated enough, they can be sent for additional evaluation to randomly selected participants.

This may be seen as a simplified *blind peer review* of contributions that makes use of voluntary participation as much as possible. Though, random selection of proposals for evaluation by a given person is functionally equivalent to random selection of participants (among currently active ones) for evaluation of a given proposal. Each reviewer thus may be considered as performing a *system-requested action*, the one of appraising a new contribution that otherwise he/she would probably never have any impulse to read.

### 3.1 Managing Public Responsiveness to System Requests

Experience from existing projects that include users' actions of appraisal (evaluation) of texts and other items and/or of their tagging shows that people are typically active enough in performing such actions on a *discretionary* (at will) basis. It is yet to be discovered how much the people involved in future large-scale eParticipation activities would be responsive to the same type of actions when they are not voluntary but system-requested. We can expect sufficient level of responsiveness, considering that possibility to participate and collaborate is the prise for itself.

If however the public will not be responsive enough, specific *incentives* could optionally be installed and managed by the support system. For example, the system could maintain for every participant two activity counters, one for his/her voluntary actions, esp. for writing his/her own contributions, another for system-requested actions performed. If the second counter is too low, new contributions from that participant will be temporarily blocked, pending a sufficient number of system-requested actions performed.

A system that randomly requires participants to do this or that in a moment that may be inopportune for a given person, may be considered by participants as too "oppressive". To avoid such an undesirable effect, the system may send its requests at any given moment to those participants only, who manifest by themselves their readiness at this moment to perform system-requested actions. In this way, the system randomly selects a contribution to be sent to a given participant, rather than a random participant to read this contribution. The result globally will be the same.

### 3.2 Two-Parameter Appraisal: How Well Presented, How Much Do I Agree

Our algorithm of clustering and ranking a large set of participants' contributions is based on their *double appraisal*, first by a few randomly selected participants (peer reviewers), then by whoever wishes to read and appraise any given contribution. Presumably, the latter ones (discretionary appraisers) would be helped by the support system when they are looking for contributions that are for them potentially the most interesting ones, and this system-made selec-tion will be made by using said algorithm, as briefly explained below; see [3] for more details.

So, every participant is expected, when reading a contribution, to appraise it on two different scales, one for the contribution's *quality*, another for the reader's *agreement* with the ideas it contains. The quality of a given text is considered to be its intrinsic characteristic, thus appealing for an objective evaluation by the reader; while the agreement is indeed a fully subjective one, as it depends on the reader's convictions, beliefs etc. Hence the two measures can be theoretically seen as independent (though, as we will see in the next section, they indeed depend on each other).

The two parameters are then exploited by the system in quite different ways. The quality grades assigned by individual appraisers are "aggregated" into the *quality rank* of every contribution; while the individually assigned agreement levels are used by the system to find contributions expressing presumably *similar or compatible ideas* or opinions. Mathematically, some *metric* is defined within the space of all contributions by pairwise comparison of agreement levels assigned by the same readers (if X agrees with both A and B, this decreases the distance between A and B; if Y agrees with A but disagrees with B, this increases the distance). Based on this metric, the system can discover *clusters* of presumably similar or compatible contributions.

Quality ranking is then performed *within every cluster*. Namely, good quality grades assigned by appraisers to a contribution will promote it towards the top of a ranked list of all contributions among those contained in a given cluster. This is indeed much more meaningful than the ordinary ranking of all contributions in a single list; namely, our method *gives voice to minorities*, whose opinions are not supported by many.

When a participant wants to read the "most interesting" contributions (or "the most representative", etc), the system will suggest him/her few highly ranked contributions within every cluster, including the smallest clusters. In this way, the participant can easily get acquainted with the whole spectrum of the currently expressed ideas.

### 3.3 System Requests for Additional Appraisals

At some point our algorithm may discover that there is not enough data to decide on a given proposal to which cluster it should go (or maybe it should start a new cluster); or to better discern the difference between two proposals or two clusters. In such cases the system may request from participants to make *additional appraisals*.

At this stage, however, all participants are not equal in their capacity to make an appraisal that will be decisive for the system. For example, if participant U hasn't yet read and appraised neither A nor B, he should do it for both, in order for the system to increase its knowledge about (dis)similarity between A and B. In contrast, if V has appraised A but not B or vice versa, then the system can ask her to appraise the other one, thus receiving a valuable information by requesting from just one requested action, rather than two in the case of U.

## 4. Rewriting Poorly Written but Agreeable Proposals

Let us now discuss in more details a very simple case of two-parameter appraisal or evaluation, where each participant is invited to characterise a proposal with just one out of three discrete values: "agree", "disagree", and "don't understand". Despite the simplicity and straightforwardness of this method, it is still a "two-parameter" one, and it can provide the system with a fairly rich information on both the set of proposals on the table, and the community of participants.

### 4.1 Agreement Depends Upon Understanding.

Although the quality in which a proposal is made, and how much does a person agree with it, are seen as independent, they are not fully so. For only inasmuch as we understand a proposal can we agree or disagree with it. We cannot agree or disagree on something we do not understand. Those two measures are a little bit like light and colour. If there is no light we cannot perceive any colour, nor does it make any sense to speak about the bandwidth of a light source when no light source is there. Hence, the brighter a source of light the better we can distinguish different colours. Here we consider only two "colours" (agree, disagree), and the brightness would represent how clearly do we understand it (see Figure 1 below; in the printed version however colours are not represented).

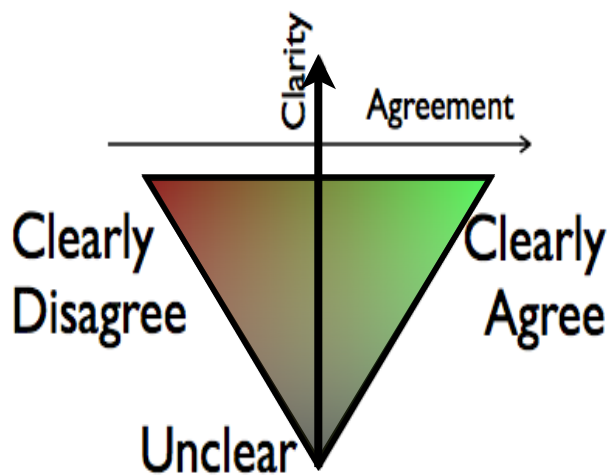

***Fig. 1.*** *Space of possible evaluations for a single proposal. Participants will be invited to evaluate every proposal with respect to how clearly do they understand it, and how much do they agree with it. The more a participant understands a proposal, the more can he/she agree or disagree with it. When a proposal is completely incomprehensible, it cannot be rated on the agreement axis.*

So the simplest form of evaluation that can still work in our system would have 3 discrete options: "agree", "disagree", and "don't understand". More continuous evaluations are also possible, always keeping in mind that if a person does not understand at all a proposal, he/she should not be allowed to evaluate how much does he/she agree with it. Also, in a continuous evaluation (or multi-grade, say with integers from -5 to +5) the span of possible agreement grades should depend on a level of understanding, so that for a medium-level understanding the uttermost agreement values, e.g. -5 or +5, should not be allowed. This quite natural restriction is graphically represented by our inverted triangle on Figure 1.

### 4.2 Using the Above Two Measures.

Once the measures have been taken and users start to evaluate the existing proposals, the system recovers a lot of useful information, much more than what can be deduced from an ordinary "linear" evaluation, even a multi-grade one.

The most obvious information that can be gathered is, first, to what extent a proposal is understandable, and second, to what extent do people agree with it. But there are also less obvious data that can be gathered, as, for example : which users are able to write proposals that are widely understandable; which users understand and agree on a specific proposal; which proposals share the same user-base (in other words, which proposals are agreed upon by the same users). Each of those characteristics can then be used in a different way.

### 4.3 Using Agreement Data for Clustering

Here we present the simplest form of our agreement-based clustering algorithm, for the above 3-value appraisal scheme. Suppose we have n proposals, and m users. Let $K_n$ be the complete weighted undirected graph with n nodes and n*(n-1)/2 edges. Let $E_{AB}$ be the edge from node A to node B; we can assign to $E_{AB}$ a *weight* that represents how many people that have voted for A have also voted for B, and how many people that have voted for B have voted for A.

Thus the weight can be defined as $W=|A \cap B|/|A \cup B|$, i.e. the cardinality of (A intersection B) divided by the cardinality of (A union B); it is a value between 0 and 1, with W=0 if no one voted for both A and B, and W=1 if everybody who voted for A voted for B and vice versa.

There are many straightforward ways to cluster the nodes in such a weighted graph; the simplest one is to delete every edge that has a weight less than x (see [5]). Clustering the nodes of a graph is a well researched area, so we can just use one of those existing methods to perform proposal clustering in a mass online deliberation (see e.g. [4]).

### 4.4 Using Quality Evaluation

While agreement evaluation can and should be used to cluster proposals, quality evaluation (i.e. in our case *clarity*, that is, how understandable they are) can be used for a different purpose. Namely, by counting the proportion of participants who didn't understood any given proposal, the system can discover which ones are not clearly written, and then invite somebody among the participants to rewrite those too obscure proposals.

We do not think it practical to suggest for an author to rewrite his/her own proposal, for we assume that each author has already done his/her best. But we also do not want to just ask a random user to rewrite a proposal. Instead, the person to rewrite a proposal that needs to be clarified should (a) understand the proposal, (b) be able to write (proposals) well, and possibly better than the original author, and (c) agree with the proposal. Although the point (c) is not strictly necessary, a person that does not agree with a proposal might not wish to spend time and effort in trying to rewrite it, and might be biased against it.

On the other side, a proposal that needs to be rewritten should (a) be widely considered incomprehensible, (b) be such that those people who do understand it would generally support it (otherwise there is no real need to clarify it), and (c) possibly be part of a relatively small cluster, so that the same ideas are not already present in some similar proposal(s).

If all that is true, then the proposal possibly has some interesting, but not clearly expressed ideas; ideas that would be lost if they are not explained better. Thanks to the above-described appraisal method, our support system will know all that information. Hence, for each proposal that needs to be rewritten, the system can find who should rewrite the proposal, and ask that person to rewrite it.

Once the new proposal (or a new version of the old proposal) has been written, the initial author can be contacted and asked if he/she agrees that the new proposal still expresses his/her ideas. If he/she agrees on that, then the new proposal can be advertised to all those people who supported the original proposal, and also to those who had declared that they did not understood it. The new proposal can be presented as "written by 1st author and 2nd author on an original idea of 1st author". Indeed the process can have several iterations and more co-authors, though probably few ideas are so complex as to require more than one rewriting.